\documentclass[10pt,twocolumn,letterpaper]{article}

\usepackage{wacv}
\usepackage{times}
\usepackage{epsfig}
\usepackage{graphicx}
\usepackage{amsmath}
\usepackage{amssymb}
\usepackage{caption}
\usepackage{subfigure}
\usepackage{dsfont}
\usepackage[pagebackref=true,breaklinks=true,letterpaper=true,colorlinks,bookmarks=false]{hyperref}

\newcommand{\norm}[1]{\left\lVert#1\right\rVert}
\newcommand{\vpara}[1]{\vspace{0.1in}\noindent\textbf{#1 }}

\wacvfinalcopy 

\def\wacvPaperID{247} 

\ifwacvfinal\fi
\begin{document}

\title{Neural Puppet: Generative Layered Cartoon Characters}

\author{Omid Poursaeed$^{1,2}$ \qquad Vladimir G. Kim$^{3}$ \qquad Eli Shechtman$^{3}$ \qquad Jun Saito$^{3}$ \qquad Serge Belongie$^{1,2}$ \\
    \newline  \\
	{$^1${Cornell University}\qquad}
    $^2${Cornell Tech}\qquad
    $^3${Adobe Research}\\	
}

\maketitle
\ifwacvfinal\thispagestyle{empty}\fi

\begin{abstract}
\vspace{-0.2cm}

We propose a learning based method for generating new animations of a cartoon character given a few example images. Our method is designed to learn from a traditionally animated sequence, where each frame is drawn by an artist, and thus the input images lack any common structure, correspondences, or labels.
We express pose changes as a deformation of a layered 2.5D template mesh, and devise a novel architecture that learns to predict mesh deformations matching the template to a target image. This enables us to extract a common low-dimensional structure from a diverse set of character poses. We combine recent advances in differentiable rendering as well as mesh-aware models to successfully align common template even if only a few character images are available during training.
In addition to coarse poses, character appearance also varies due to shading, out-of-plane motions, and artistic effects. We capture these subtle changes by applying an image translation network to refine the mesh rendering, providing an end-to-end model to generate new animations of a character with high visual quality. 
We demonstrate that our generative model can be used to synthesize in-between frames and to create data-driven deformation. Our template fitting procedure outperforms state-of-the-art generic techniques for detecting image correspondences.
\end{abstract}

\begin{figure}[h]
\begin{center}
   \includegraphics[width=0.93\linewidth]{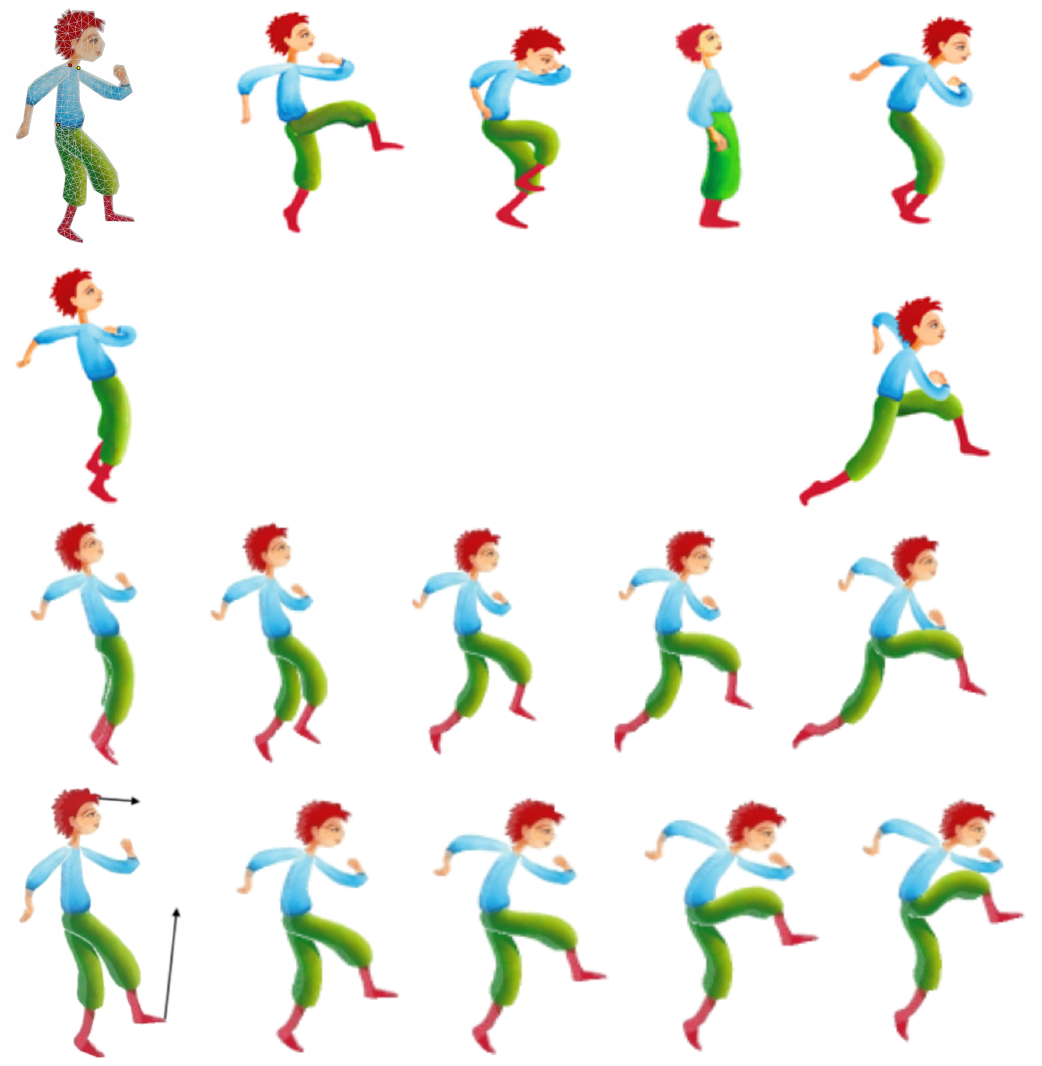}
\end{center}
\vspace{-0.65cm}
   \caption{The user provides a template mesh and a set of unlabeled images (first row). The model learns to generate inbetween frames (row 2 and 3) and constrained deformations (row 4). }\label{fig:teaser}
\vspace{-0.75cm}
\end{figure}

\vspace{-0.5cm}
\section{Introduction}

Traditional character animation is a tedious process that requires artists meticulously drawing every frame of a motion sequence. After observing a few such sequences, a human can easily imagine what a character might look in other poses, however, making these inferences is difficult for learning algorithms. The main challenge is that the input images commonly exhibit substantial variations in appearance due to articulations, artistic effects, and viewpoint changes, significantly complicating the extraction of the underlying character structure. 
In the space of natural images, one can rely on extensive annotations~\cite{alp2018densepose} or vast amount of data~\cite{shu2018deforming} to extract the common structure. Unfortunately, this approach is not feasible for cartoon characters, since their topology, geometry, and drawing style is far less consistent than that of natural images of human bodies or faces. 

To tackle this challenge, we propose a method that learns to generate novel character appearances from a small number of examples by relying on additional user input: a deformable puppet template. We assume that all character poses can be generated by warping the deformable template, and thus develop a deformation network that encodes an image and decodes deformation parameters of the template. These parameters are further used in a differentiable rendering layer that is expected to render an image that matches the input frame. Reconstruction loss can be back-propagated through all stages, enabling us to learn how to register the template with all of the training frames. While the resulting renderings already yield plausible poses, they fall short of artist-generated images since they only warp a single reference, and do not capture slight appearance variations due to shading and artistic effects. To further improve visual quality of our results, we use an image translation network that synthesizes the final appearance.  

While our method is not constrained to a particular choice for the deformable puppet model, we chose a layered 2.5D deformable model that is commonly used in academic~\cite{Correa:1998:TMF} and industrial~\cite{Ch:2019} applications. This model matches many traditional hand-drawn animation styles, and makes it significantly easier for the user to produce the template relative to 3D modeling that requires extensive expertise. To generate the puppet, the user has to select a single frame and segment the foreground character into constituent body parts, which can be further converted into meshes using standard triangulation tools~\cite{shewchuk1996triangle}.  
%

We evaluate our method on 
animations of six characters with 70\%-30\% train-test split. 
First, we evaluate how well our model can reconstruct the input frame and demonstrate that it produces more accurate results than state-of-the-art optical flow and auto-encoder techniques. Second, we evaluate the quality of correspondences estimated via the registered templates, and demonstrate improvement over image correspondence methods. Finally, we show that our model can be used for data-driven animation, where synthesized animation frames are guided by character appearances observed at training time. We build prototype applications for synthesizing in-between frames and animating by user-guided deformation where our model constrains new images to lie on a learned manifold of plausible character deformations. We show that the data-driven approach yields more realistic poses that better match to original artist drawings than traditional energy-based optimization techniques used in computer graphics.

 \vspace{-0.1cm}
 \section{Related Work}
 \vspace{-0.2cm}
\vpara{Deep Generative Models. } Several successful paradigms of deep generative models have emerged recently, including the auto-regressive models \cite{gregor2013deep,oord2016pixel, van2016conditional}, Variational Auto-encoders (VAEs) \cite{kingma2013auto, kingma2014semi, rezende2014stochastic}, and Generative Adversarial Networks (GANs) \cite{goodfellow2014generative, radford2015unsupervised, salimans2016improved, huang2017stacked, arjovsky2017wasserstein,kiapour2019generating}. Deep generative models have been applied to image-to-image translation \cite{isola2017image, zhu2017unpaired, zhu2016generative}, image superresolution~\cite{ledig2017photo}, learning from synthetic data~\cite{bousmalis2017unsupervised,shrivastava2017learning}, generating adversarial images~\cite{poursaeed2018generative,poursaeed2019fine}, and
synthesizing 3D volumes~\cite{tatarchenko2017octree, soltani2017synthesizing}.
These techniques usually make no assumptions about the structure of the training data and synthesize pixels (or voxels) directly. This makes them very versatile and appealing when a large number of examples are available. Since these data might not be available in some domains, such as 3D modeling or character animation, several techniques leverage additional structural priors to train deep models with less training data. 


\vpara{Learning to Generate with Deformable Templates. }
Deformable templates have been used for decades to address analysis and synthesis problems~\cite{amit1991structural,blanz2003face,allen2003space,allen2006learning,loper2015smpl,zuffi2015stitched}. 
Synthesis algorithms typically assume that multiple deformations of the same template (e.g., a mesh of the same character in various poses) is provided during training. Generative models, such as variational auto-encoders directly operate on vertex coordinates to encode and generate plausible deformations from these examples~\cite{Tan_2018_CVPR,kostrikov2018surface,litany2017deformable}. Alternative representations, such as multi-resolution meshes~\cite{ranjan2018generating}, single-chart UV~\cite{bagautdinov2018modeling}, or multi-chart UV~\cite{Hamu2018MultichartGS} is used for higher resolution meshes. 
This approaches are limited to cases when all of the template parameters are known for all training examples, and thus cannot be trained or make inferences over raw unlabeled data.

Some recent work suggests that 
neural networks can jointly learn the template parameterization and optimize for the alignment between the template and a 3D shape~\cite{groueix2018shape} or 2D images~\cite{kanazawa2018learning,henderson2018learning,venkatdeep}. While these models can make inferences over unlabeled data, they are trained on a large number of examples with rich labels, such as dense or sparse correspondences defined for all pairs of training examples.
%

To address this limitation, a recent work on deforming auto-encoders provides an approach for unsupervised group-wise image alignment of related images (e.g. human faces) \cite{shu2018deforming}. They disentangle shape and appearance in latent space, by predicting a warp of a learned template image as well as its texture transformations to match every target image. Since their warp is defined over the regular 2D lattice, their method is not suitable for strong articulations. 
Thus, to handle complex articulated characters and strong motions, we leverage an user-provided template and rely on regularization terms that leverage the rigging as well as mesh structure. Instead of synthesizing appearance in a common texture space, we do the final image translation pass that enables us to recover from warping artifacts and capture effects beyond texture variations, such as out-of-plane motions. 
%


\vpara{Mesh-based models for Character Animation.}
Many techniques have been developed to simplify the production of traditional hand-drawn animation using computers~\cite{catmull1978problems,di2001automatic}. Mesh deformation techniques enable novice users to create animations by manipulating a small set of control points of a single puppet~\cite{sorkine2007rigid,jacobson2011biharmonic}. 
To avoid overly-synthetic appearance, one can further stylize these deformations by leveraging multiple co-registered examples to guide the deformation~\cite{wampler2016fast}, and final appearance synthesis~\cite{dvorovzvnak2017example,dvorovznak2018toonsynth}. These methods, however, require artist to provide the input in a particular format, and if this input is not available rely on image-based correspondence techniques~\cite{bregler2002mocaptoons,sykora2009rigid,choy2016universal,Fan18,sun2018pwc} to register the input. 
Our deformable puppet model relies on a layered mesh representation~\cite{dvorovznak2018toonsynth} and mesh regularization terms~\cite{sorkine2007rigid} used in these optimization-based workflows. Our method jointly learns the puppet deformation model as it registers the input data to the template, and thus yields more accurate correspondences than state-of-the-art flow-based approach~\cite{sun2018pwc} and state-of-the-art feature-based method trained on natural images~\cite{choy2016universal}.

\vspace{-0.1cm}
\section{Approach}
\vspace{-0.1cm}

Our goal is to learn a deformable model for a cartoon character given an unlabeled collection of images. First, the user creates a layered deformable template puppet by segmenting one reference frame. We then train a two-stage neural network that first fits this deformable puppet to every frame of the input sequence by learning how to warp a puppet to reconstruct the appearance of the input, and second, it refines the rendering of deformed puppet to account for texture variations and motions that cannot be expressed with a 2D warp. 


\subsection{A Layered Deformable Puppet}
The puppet geometry is represented with a few triangular meshes ordered into layers and connected at hinge joints. For simplicity, we denote them as one mesh with vertices $V$ and faces $F$, where every layer is a separate connected component of the mesh. Joints are represented as $\{(p_i, q_i)\}$ pairs, connecting vertices between some of these components. The puppet appearance is captured as texture image $I^{uv}$, which aligns to some rest pose $\hat{V}$. New character poses can be generated by modifying vertex coordinates, and the final appearance can be synthesized by warping the original texture according to mesh deformation. 

\begin{figure}[t]
\begin{center}
    \subfigure[]{
    \includegraphics[width=0.25\linewidth]{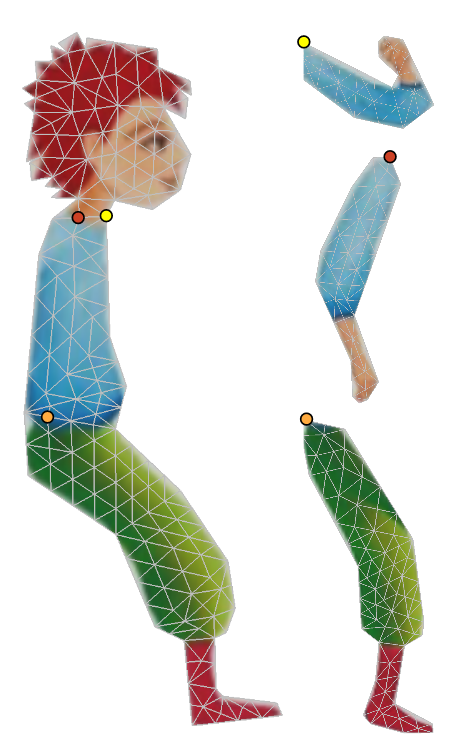}}
    \subfigure[]{
   \includegraphics[width=0.23\linewidth]{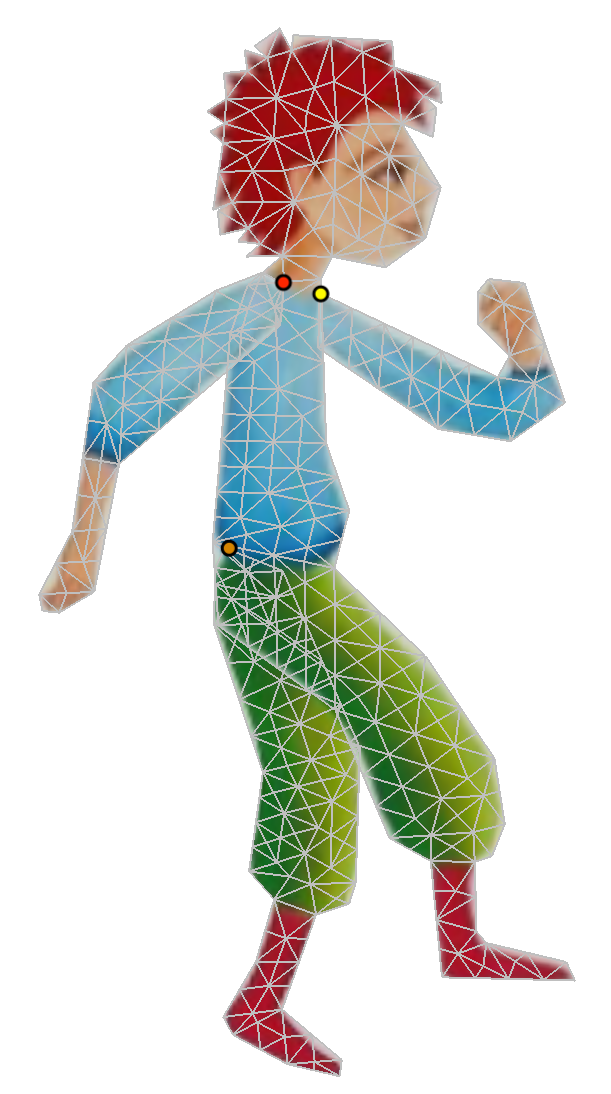}}
\end{center}
\vspace{-0.65cm}
   \caption{
   Deformable Puppet. a) For each body part a separate mesh is created, and joints (shown with circles) are specified.  
   b) The meshes are combined into a single mesh. 
   The UV-image of the final mesh consists of translated versions of separate texture maps. 
   }\label{fig:layering}
   \vspace{-0.65cm}
\end{figure}

Unlike 3D modeling, even inexperienced users can create the layered 2D puppet. First, one selects a reference frame and provides the outline for different body parts and prescribes the part ordering. We then use standard triangulation algorithm~\cite{shewchuk1996triangle} to generate a mesh for each part, and create a hinge joint at the centroid of overlapping area of two parts. 
We can further run midpoint mesh subdivision to get a finer mesh that can model more subtle deformations. Figure \ref{fig:layering} illustrates a deformable puppet. 

\begin{figure*}[t]
\begin{center}
   \includegraphics[width=0.92\linewidth]{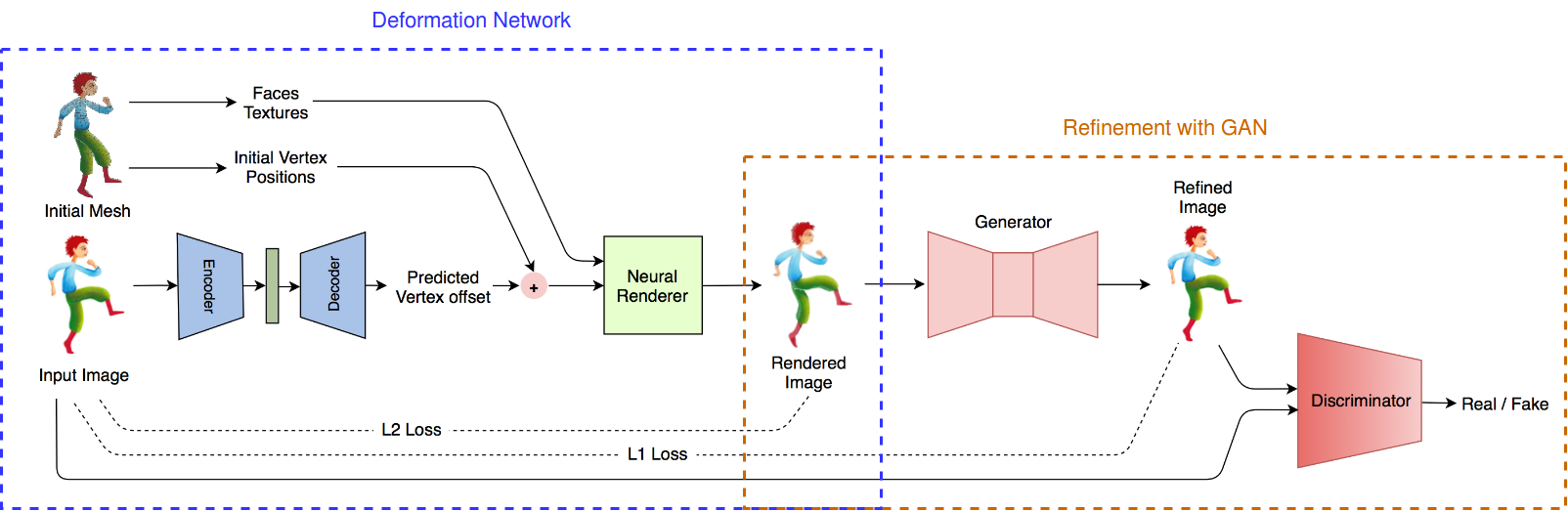}
\end{center}
\vspace{-0.5cm}
   \caption{Training Architecture. An encoder-decoder network learns the mesh deformation and a conditional Generative Adversarial Network refines the rendered image to capture texture variations. }\label{fig:architecture-gan}
\vspace{-0.55cm}
\end{figure*}

\subsection{Deformation Network}
After obtaining the template, we aim to learn how to deform it to match a target image of the character in a new pose. Figure \ref{fig:architecture-gan} illustrates our architecture. 
%
%
%
The inputs to the Deformation Network 
are the initial mesh and a target image of the character in a new pose. An encoder-decoder network takes the target image, encodes it to a 512-dimensional bottleneck via three convolutional filters\footnote{Kernel size=5, padding=2, stride=2} and three fully connected layers, and then decodes it to per-vertex position offsets and a global bias via three fully connected layers. 
Deformation network learns to recognize the pose in the input image and then infer appropriate template deformation to reproduce the pose.  We assume the connectivity of vertices and the textures remain the same compared to the template. Hence, we pass the faces and textures of the initial mesh in tandem with the predicted vertex positions to a differentiable renderer $R$. The rendered image is then compared to the input image using $L_2$ reconstruction loss:
\begin{equation}
    L_{rec} = \norm{x - R(V_{pred}, F, I^{uv})}^2
\end{equation}
in which $x$ represents the input image. We use the Neural Mesh Renderer \cite{kato2018neural} as our differentiable renderer, since it can be easily integrated into neural network architectures. {Note that the number of vertices can vary for different characters as we train a separate model for each character.} 


\vpara{Regularization.} 
The model trained with only the reconstruction loss does not preserve the structure of the initial mesh, and the network may generate large deformations to favor local consistency. 
In order to prevent this, we use the ARAP regularization energy \cite{sorkine2007rigid} which penalizes deviations of per-cell transformations from rigidity:
\vspace{-0.25cm}
\begin{equation}
    L_{reg} = \sum_{i=1}^{|V|} \sum_{j\in \mathcal{N}_i} w_{ij} \norm{(\hat{v}_i-\hat{v}_j) - R_i (v_i - v_j)}^2
\end{equation}
in which $v_i$ and $\hat{v}_i$ are coordinates of vertex $i$ before and after deformation, $\mathcal{N}_i$ denotes neighboring vertices of vertex $i$, $w_{ij}$ are cotangent weights and $R_i$ is the optimal rotation matrix as discussed in \cite{sorkine2007rigid}. 

\vpara{Joints Loss.} If we do not constrain vertices of the layered mesh, different limbs can move away from each other, resulting in unrealistic outputs. In order to prevent this, we specify `joints' $(p_i, q_i), i=1,\ldots,n$ as pairs of vertices in different layers that must remain close to each other after deformation. We manually specify the joints
, and penalize the sum of distances between vertices in each joint: 
\vspace{-0.2cm}
\begin{equation}
\vspace{-0.25cm}
    L_{joints}=\sum_{i=1}^{n}\norm{p_i - q_i}^2
\end{equation}

Our final loss for training the Deformation Network is a linear combination of the 
aforementioned losses:
\begin{equation}\label{eq:L_total}
    L_{total} = L_{rec} + \lambda_1 \cdot L_{reg} +  \lambda_2 \cdot L_{joints}
\end{equation}
We use $\lambda_1=2500$ and $\lambda_2=10^{4}$ in the experiments.  

\subsection{Appearance Refinement Network}
While articulations can be mostly captured by deformation network, some appearance variations such as artistic stylizations, shading effects, and out-of-plane motions cannot be generated by warping a single reference. To address this limitation, we propose an appearance refinement network that processes the image produced by rendering the deformed puppet. Our architecture and training procedure is similar to conditional Generative Adversarial Network (cGAN) approach that is widely used in various domains~\cite{mirza2014conditional,yoo2016pixel,isola2017image}. 
%
The corresponding architecture is shown in Figure \ref{fig:architecture-gan}. 
The generator refines the rendered image to look more natural and more similar to the input image. The discriminator tries to distinguish between input frames of character poses and generated images. These two networks are then trained in an adversarial setting \cite{goodfellow2014generative}, where we use pix2pix architecture~\cite{isola2017image} and Wasserstein GAN with Gradient Penalty for adversarial loss~\cite{arjovsky2017wasserstein, gulrajani2017improved}: 
\vspace{-0.1cm}
\begin{equation}
    L_G = -\mathbb{E} \big[D(G(x_\text{rend}))\big] + \gamma_1 \norm{G(x_\text{rend}) - x_\text{input}}_1
\end{equation}
And the discriminator's loss is: 
\vspace{-0.1cm}
\begin{multline}
    L_D = \mathbb{E}\big[D(G(x_\text{rend}))\big] - \mathbb{E}\big[D(x_\text{real})\big] + \\ 
    \gamma_2 \ \mathbb{E}\big[ (\norm{\nabla_{\hat{x}} D(\hat{x})}_2 - 1)^2\big] 
\end{multline}
in which $D( \cdot )$ and $G( \cdot )$ are the discriminator and the generator, $\gamma_1,\gamma_2\in \mathbb{R}$ are  weights, $x_\text{input}$ and $x_\text{rend}$ are the input and rendered images, $x_\text{real}$ is an image sampled from the training set, and $\hat{x} = \epsilon {\ } G(x_\text{rend}) + (1 - \epsilon) {\ } x_\text{real} $
with $\epsilon$ uniformly sampled from the [0, 1] range. The cGAN is trained independently after training the Deformation Network as this results in more stable training. {Note that one could also use deformed results directly without the refinement network. }
%


\vspace{-0.2cm}
\section{Results and Applications} 
%
\begin{figure*}
\begin{center}
   \includegraphics[width=0.95\linewidth]{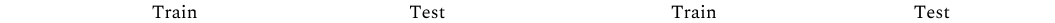}
   \includegraphics[width=0.94\linewidth]{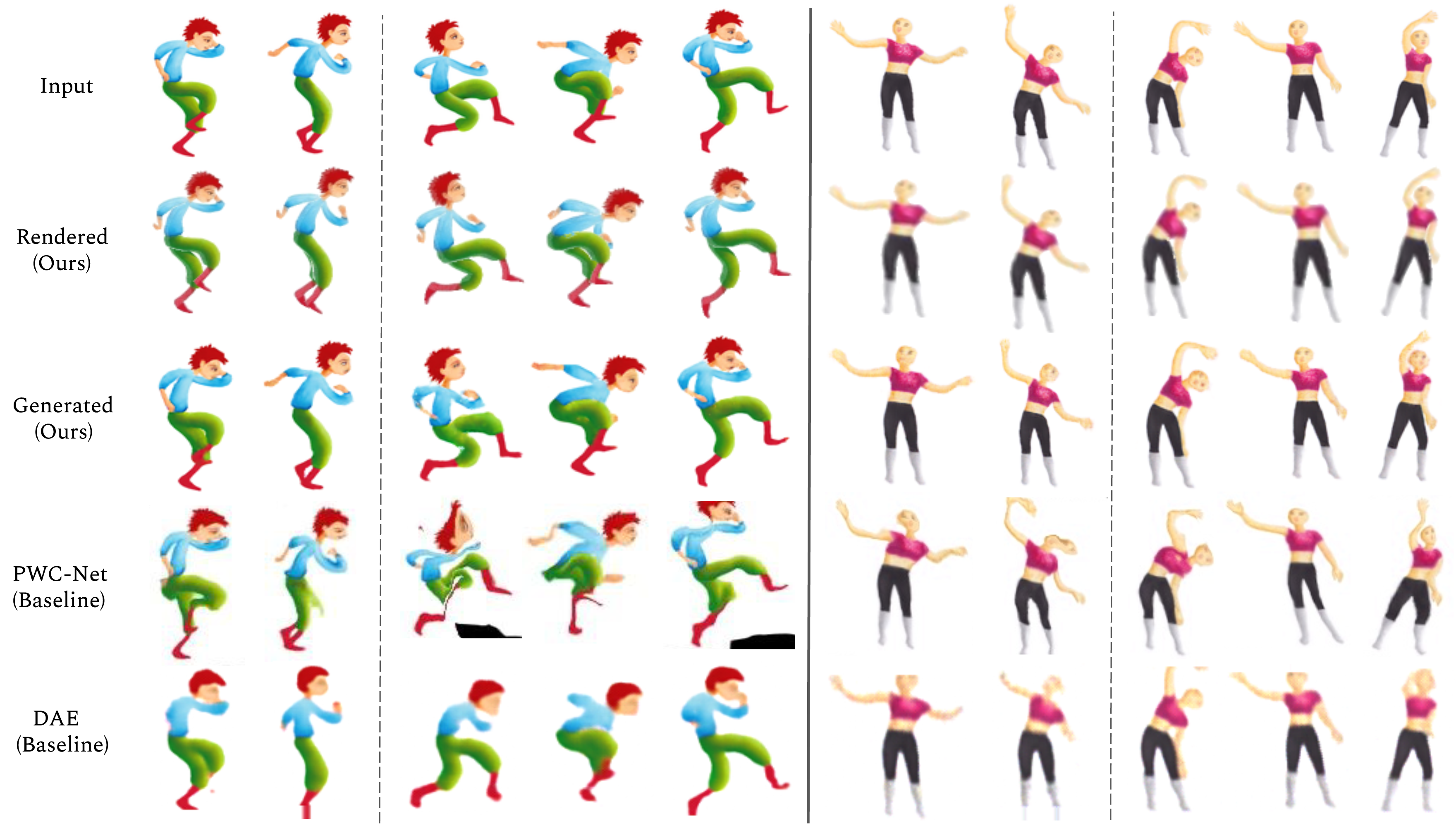}
     \includegraphics[width=0.94\linewidth]{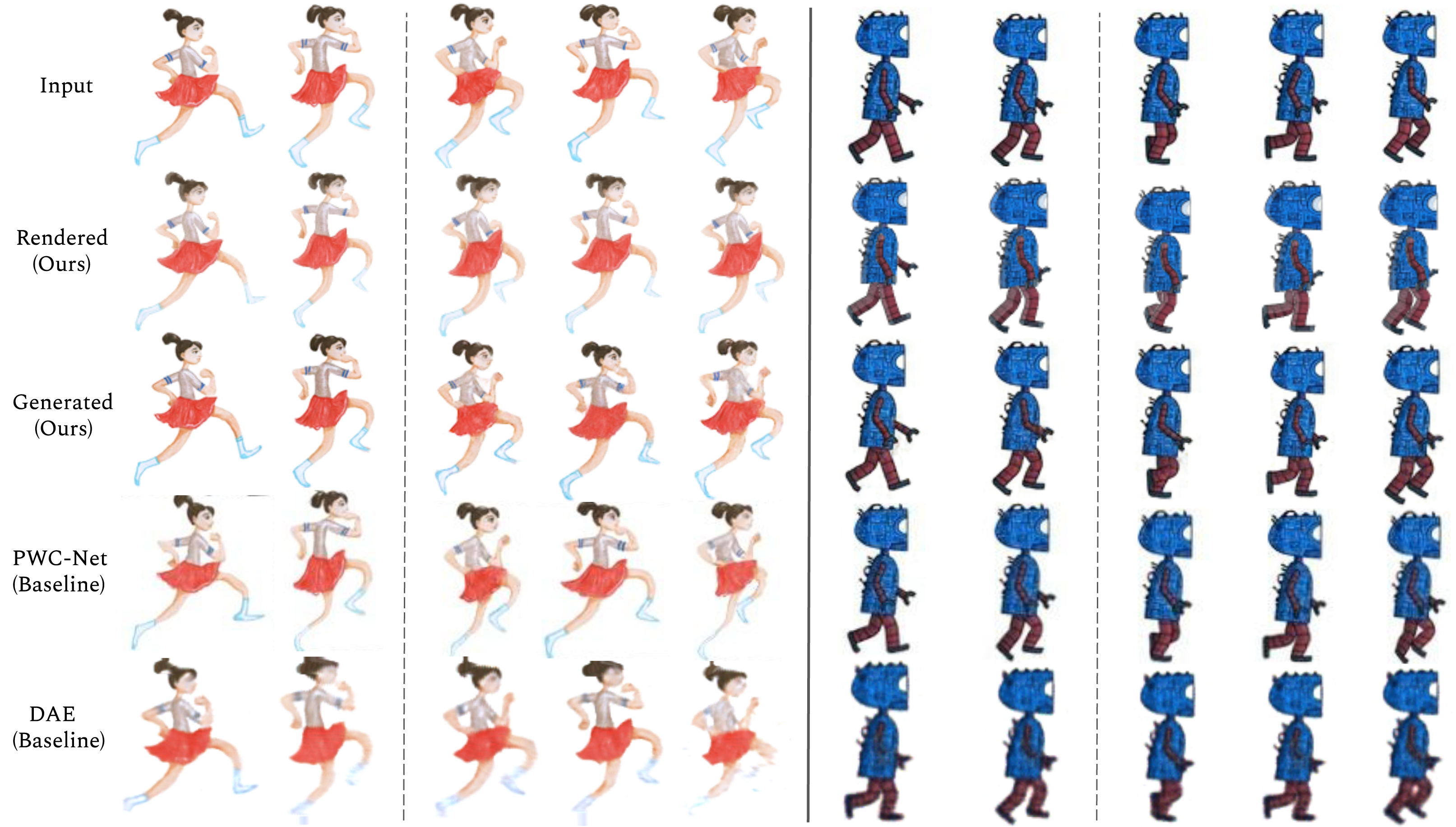}
\end{center}
\vspace{-0.5cm}
   \caption{Input images, our rendered and final results, followed by results obtained with PWC-Net~\cite{sun2018pwc} and DAE~\cite{shu2018deforming} (input images for the first three characters are drawn by \textcopyright{Zuzana Studen\'{a}}. The fourth character \textcopyright{Adobe Character Animator}).}\label{fig:results}
   \vspace{-0.55cm}
\end{figure*} 

We evaluate how well our method captures (i.e., encodes and reconstructs) character appearance. We use six character animation sequences from various public sources. Figure~\ref{fig:results} shows some qualitative results where for each input image we demonstrate output of the deformation network (\emph{rendered}) and the final synthesized appearance (\emph{generated}). The first three characters are from Dvoroznak et al.~\cite{dvorovznak2018toonsynth} with 1280/547, 230/92 and 60/23 train/test images respectively. The last character (robot) is obtained from Adobe Character Animator \cite{Ch:2019} with 22/9 train/test images. Other characters and their details are given in the supplementary material.  
The \emph{rendered} result is produced by warping a reference puppet, and thus it has fewer degrees of freedom (e.g., it cannot change texture or capture out-of-plane motions). However, it still provides fairly accurate reconstruction even for very strong motions, suggesting that our layered puppet model makes it easier to account for significant character articulations, and that our image encoding can successfully predict these articulations even though it was trained without strong supervision. 
Our refinement network does not have any information from the original image other than the re-rendered pose, but, evidently, adversarial loss provides sufficient guidance to improve the final appearance and match the artist-drawn input. 
To confirm that these observations hold over all characters and frames, we report $L_2$ distance between the target and generated images in Table~\ref{tab:pwc}. See supplemental material for more examples. 
%

We compare our method to alternative techniques for re-synthesizing novel frames (also in Figure~\ref{fig:results}). One can use optical flow method, such as PWC-Net \cite{sun2018pwc} to predict a warping of a reference frame (we use the frame that was used to create the puppet) to the target image. This method was trained on real motions in videos of natural scenes and tends to introduce strong distortion artifacts when matching large articulations in our cartoon data. 
Various autoencoder approaches can also be used to encode and reconstruct appearance. We compare to a state-of-the-art approach that uses deforming auto-encoders~\cite{shu2018deforming} to disentangle deformation and appearance variations by separately predicting a warp field and a texture. This approach does not decode character-specific structure (the warp field is defined over a regular grid), and thus also tends to fail at larger articulations. Another limitation is that it controls appearance by altering a pre-warped texture, and thus cannot correct for any distortion artifacts introduced by the warp. 
Both methods have larger reconstruction loss in comparison to our rendered as well as final results (see Table~\ref{tab:pwc}).

  \begin{table}
\begin{center}
\begin{tabular}{ |c|c|c|c|c|c| }
\hline
 & Char1 & Char2 & Char3 & Char4 & Avg \\ 
  \hline
  Rendered &  819.8 & 732.7  & 764.1 & 738.9 & 776.1 \\
 \hline
 Generated & \textbf{710.0} & \textbf{670.5} & \textbf{691.7}  & \textbf{659.2} & \textbf{695.3} \\
 \hline
 PWC-Net & 1030.4 & 1016.1 & 918.3 & 734.6 & 937.1 \\
\hline 
DAE  & 1038.3 & 1007.2 & 974.8 & 795.1 & 981.6 \\
\hline 
\end{tabular}
\end{center}
\vspace{-0.45cm}
\caption{Average $L_2$ distance to the target images from the test set. Rendered and generated images from our method are compared with PWC-Net \cite{sun2018pwc} and Deforming Auto-encoders \cite{shu2018deforming}. 
The last column shows the average distance across six different characters. 
} \label{tab:pwc}
\vspace{-0.5cm}
\end{table}

One of the key advantages of our method is that it predicts deformation parameters, and thus can retain resolution of the artist-drawn image. To illustrate this, we render the output of our method as $1024 \times 1024$ images using vanilla OpenGL renderer in the supplementary material. 
The final conditional generation step can also be trained on high-resolution images.


\subsection{Inbetweening}
\vspace{-0.1cm}
In traditional animation, a highly-skilled artist creates a few sparse keyframes and then in-betweening artists draw the other frames to create the entire sequence. Various computational tools have been proposed to automate the second step~\cite{kort2002computer,sykora2009rigid,baxter2009n,whited2010betweenit}, but these methods typically use hand-crafted energies to ensure that intermediate frames look plausible, and rely on the input data to be provided in a particular format (e.g., deformations of the same puppet). Our method can be directly used to interpolate between two raw images, and our interpolation falls on the manifold of deformations learned from training data, thus generating in-betweens that look similar to the input sequence. 

Given two images $x_1$ and $x_2$ we use the encoder in deformation network to obtain their features, $z_i=E(x_i)$. We then linearly interpolate between $z_1$ and $z_2$ with uniform step size, and for each in-between feature $z$ we apply the rest of our network to synthesize the final appearance. The resulting interpolations are shown in Figure~\ref{fig:in-betweening} and in supplemental video. 
The output images smoothly interpolate the motion, while mimicking poses observed in training data. This suggests that the learned manifold is smooth and can be used directly for example-driven animation. We further confirm that our method generalizes beyond training data by showing nearest training neighbor to the generated image (using Euclidean distance as the metric). 

%
%

\begin{figure*}[t]
\begin{center}
      \includegraphics[width=0.92\linewidth]{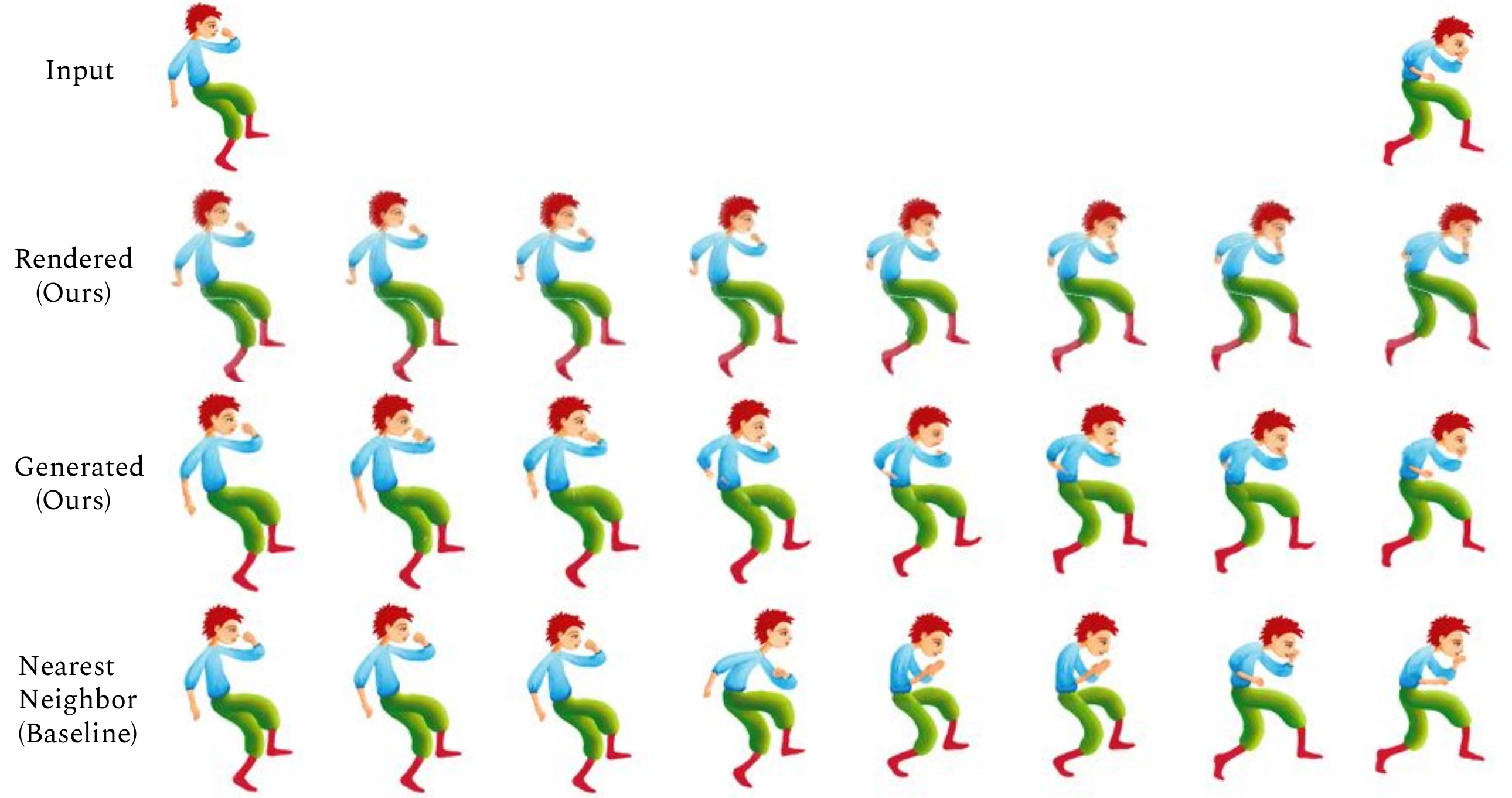}
\end{center}
\vspace{-0.6cm}
   \caption{Inbetweening results. Two given images (first row) are encoded. The resulting latent vectors are linearly interpolated yielding the rendered and generated (final) images. For each generated image, the corresponding Nearest Neighbor image from the training set is retrieved.}\label{fig:in-betweening}
\vspace{-0.35cm}
\end{figure*}

  \begin{figure*}[t]
\begin{center}
  \includegraphics[width=0.92\linewidth]{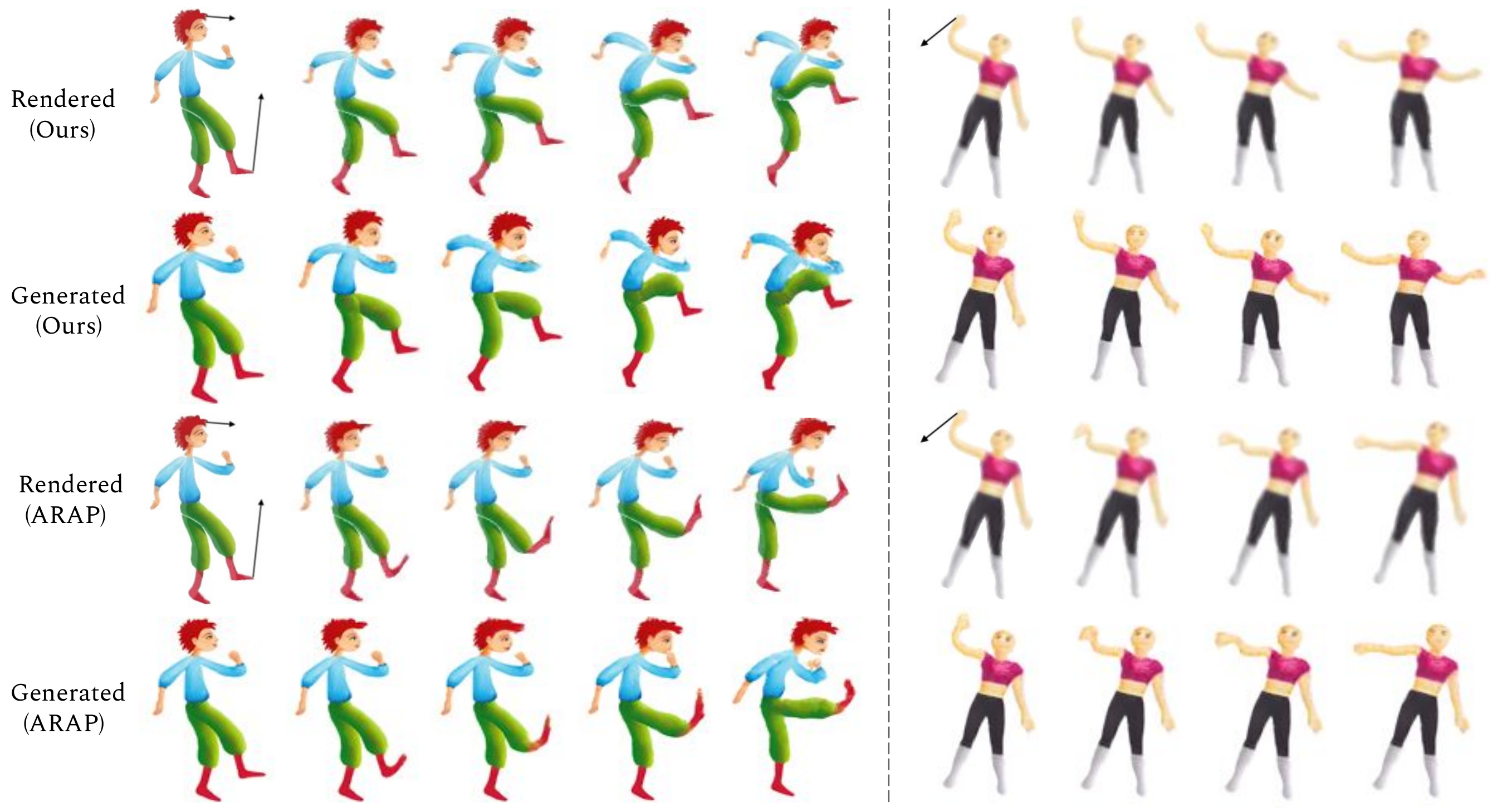}
\end{center}
\vspace{-0.65cm}
   \caption{User-constrained deformation. Given the starting vertex and the desired location (shown with the arrow), the model learns a plausible deformation to satisfy the user constraint. Our approach of searching for an optimal latent vector achieves global shape consistency, while optimizing directly on vertex positions only preserves local rigidity.}\label{fig:user-constrained}
\vspace{-0.55cm}
\end{figure*}


\subsection{User-constrained Deformation}
Animations created with software assistance commonly rely on deforming a puppet template to target poses. These deformations are typically defined by local optima with respect to user-prescribed constraints (i.e., target motions) and some hand-crafted energies such as rigidity or elasticity~\cite{sorkine2007rigid,levi2015smooth,chao2010simple}. This is equivalent to deciding on what kind of physical material the character is made of (e.g., rubber, paper), and then trying to mimic various deformations of that material without accounting for artistic stylizations and bio-mechanical priors used by professional animators.
While some approaches allow transferring these effects from stylized animations~\cite{dvorovznak2018toonsynth}, they require artist to provide consistently-segmented and densely annotated frames aligned to some reference skeleton motion. 
Our model does not rely on any annotations and we can directly use our learned manifold to find appropriate deformations that satisfy user constraints.

Given the input image of a character $x$, the user clicks on any number of control points $\{p_i\}$ and prescribes their desired target positions $\{p_i^\text{trg}\}$. Our system then produces the image $x'$ that satisfies the user constraints, while adhering to the learned manifold of plausible deformations. First, we use the deformation network to estimate vertex parameters to match our puppet to the input image: $\mathbf{v}=D(E(x))$ (where $E(\cdot)$ is the encoder and $D(\cdot)$ is the decoder in Figure~\ref{fig:architecture-gan}). We observe that each user-selected control point $p_i$ can now be found on the surface of the puppet mesh. One can express its position as a linear combination of mesh vertices,  $p_i(\mathbf{v})$, where we use barycentric coordinates of the triangle that encloses $p_i$. The user constrained can be expressed as an energy, penalizing distance to the target:
\vspace{-0.15cm}
\begin{equation}
    L_{user} = \sum_i \norm{p_i(\mathbf{v}) - p^\text{trg}_i}^2
\end{equation}
\vspace{-0.45cm}

For the deformation to look plausible, we also include the regularization terms:
\vspace{-0.15cm}
\begin{equation}
    L_{deform} = L_{user} + \alpha_1 \cdot L_{reg} + \alpha_2 \cdot L_{joints}
\end{equation}
We use $\alpha_1 = 25$,  $\alpha_2 = 50$ in the experiments. 

\begin{figure*}[t!]
\begin{center}
   \includegraphics[width=0.92\linewidth]{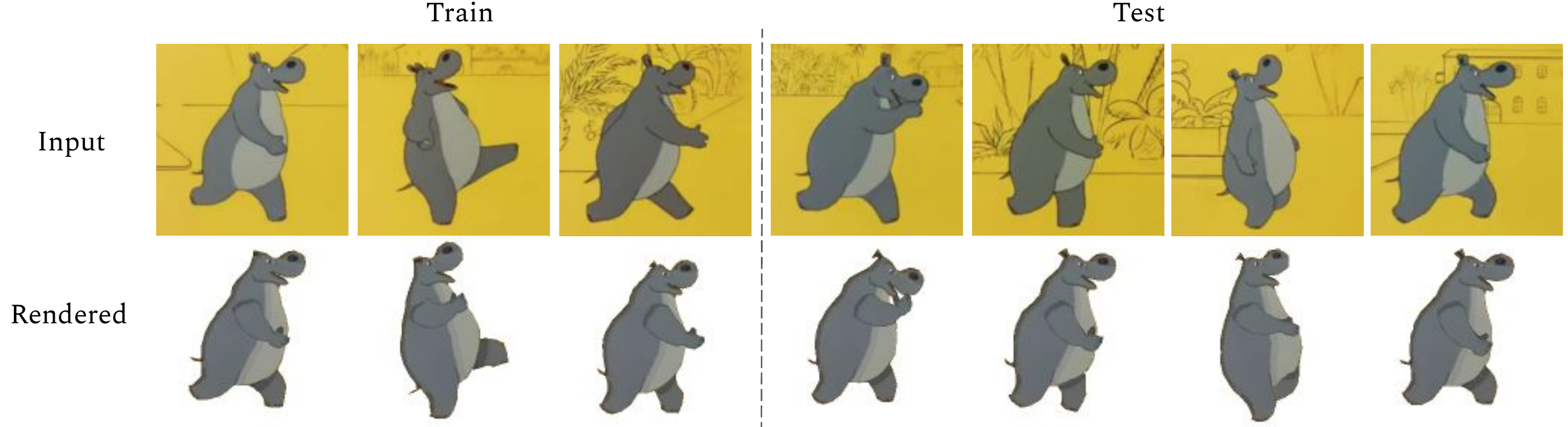}
\end{center}
\vspace{-0.6cm}
   \caption{Characters in the wild. 
   The model learns both the outline and the pose of the character (input frames and the character \textcopyright{Soyuzmultfilm}).}\label{fig:results-bg}
 \vspace{-0.45cm}
\end{figure*}

Since our entire deformation network is differentiable, we can propagate the gradient of this loss function to the embedding of the original image $z_0=E(x)$ and use gradient descent to find $z$ that optimizes $L_{deform}$:
\vspace{-0.15cm}
\begin{equation}
    z \longleftarrow z - \eta \nabla_{z}L_{deform}
\end{equation}
%
where $\eta = 3 \times 10^{-4}$ is the step size. 
%
In practice, a few (2 to 5) iterations of gradient descent suffice to obtain satisfactory results, enabling fast, interactive manipulation of the mesh by the user (in the order of seconds).  

%
%



The resulting latent vector is then passed to the decoder, the renderer and the refinement network. 
Figure \ref{fig:user-constrained} (left) illustrates the results of this user-constrained deformation.
As we observe, deformations look plausible and satisfy the user constraints. They also show global consistency; for instance, as we move one of the legs to satisfy the user constraint, the torso and the other leg also move in a consistent manner. This is due to the fact that the latent space encodes high-level information about the character's poses, and it learns that specific poses of torso are likely to co-occur with specific poses of legs, as defined by the animator.  

We compare our method with optimizing directly in the vertex space using the regularization terms only (Figure \ref{fig:user-constrained}, right). This approach does not use the latent representation, and thus does not leverage the training data. It is similar to traditional energy-based approaches, where better energy models might yield smoother deformation, but would not enforce long-range relation between leg and torso motions. 

\begin{table}
\begin{center}
\begin{tabular}{ |c|c|c|c|c| }
\hline
 & $\alpha=0.1$ & $\alpha=0.05$ & $\alpha=0.025$ \\ 
 \hline
 Ours & \textbf{67.18} & \textbf{46.39} & \textbf{24.17} \\
 \hline
 UCN & 67.07 & 43.84 & 21.50 \\
 \hline
  PWC-Net & 62.92 & 40.74 & 18.47 \\
 \hline
\end{tabular}
\end{center}
\vspace{-0.45cm}
\caption{Correspondence estimation results using PCK as the metric. Results are averaged across six characters.  
} \label{tab:correspondence}
\vspace{-0.65cm}
\end{table}

\vspace{-0.1cm}
\subsection{Correspondence Estimation}
\vspace{-0.1cm}
Many video editing applications require inter-frame correspondence. While many algorithms have been proposed to address this goal for natural videos \cite{choy2016universal,liu2011sift,kim2013deformable,yang2014daisy,revaud2016deepmatching}, they are typically not suitable for cartoon data, as it usually lacks texture and exhibits strong expressive articulations. Since our Deformation Network fits the same template to every frame, we can estimate correspondences between any pair of frames via the template. 
Table \ref{tab:correspondence} compares correspondences from our method with those obtained from a recent flow-based approach (PWC-Net \cite{sun2018pwc}), and with a supervised correspondence method, Universal Correspondence Networks (UCN) \cite{choy2016universal}. 
 We use the Percentage of Correct Keypoints (PCK)  
 as the evaluation metric. 
Given a threshold $\alpha\in(0, 1)$, the correspondence is classified as correct if the predicted point lies within Euclidean distance $\alpha\cdot L$ of the ground-truth, in which $L=\text{max}(\text{width, \text{height}})$ is the image size. Results are averaged across pairs of images in the test set for six different characters. 
Our method outperforms UCN and PWC-Net in all cases, since it has a better model for underlying character structure. 
Note that our method requires a single segmented frame, while UCN is trained with ground truth key-point correspondences, and PWC-Net is supervised with ground truth flow. 


\subsection{Characters in the Wild}
We also extend our approach to TV cartoon characters ``in the wild''. The main challenge posed by these data is that the character might only be a small element of a complex scene. Given a raw video\footnote{e.g. https://www.youtube.com/watch?v=hYCbxtzOfL8}, we first use the standard tracking tool in Adobe After Effects~\cite{AE:2019} to extract per-frame bounding boxes that encloses the character. Now we can use our architecture to only analyze the character appearance. However, since it still appears over a complex background, we modify our reconstruction loss to be computed only over the rendered character:
\vspace{-0.2cm}
\begin{equation}
    L^\text{masked}_\text{rec} = \frac{\norm{x \odot{R(V_{pred}, F, I^\mathds{1}) } - {R(V_{pred}, F,  {I^{uv}})}}^2}{\sum R(V_{pred}, F, I^\mathds{1})}
\end{equation}
where $x$ is the input image, $\odot$ is the Hadamard product, and $R(V_{pred}, F, I^\mathds{1})$ is a mask, produced by rendering a mesh with 1-valued (white) texture over a 0-valued (black) background. By applying this mask, we compare only the relevant regions of input and rendered images, i.e. the foreground character. The term in the denominator normalizes the loss by the total character area. To further avoid shrinking or expansion of the character (which could be driven by a partial match), we add an area loss penalty:
\vspace{-0.25cm}
\begin{equation}\label{eq:L_area}
    L_{area} =\Big{|}{ \sum R(V_{pred}, F, I^\mathds{1}) - \sum R(V_{init}, F, I^\mathds{1})\Big{|} }^2
\end{equation}
Our final loss is defined similarly to Equation~\ref{eq:L_total} but uses the masked reconstruction loss $L^\text{masked}_\text{rec}$ and adds $L_{area}$ loss with weight  $2\times 10^{-3}$ ($L_{reg}$ and $L_{joints}$ are included with original weights).
We present our results in Figure~\ref{fig:results-bg}, demonstrating that our framework can be used to capture character appearance in the wild. 

\vspace{-0.1cm}
\section{Conclusion and Future Work}
\vspace{-0.1cm}
We present novel neural network architectures for learning to register deformable mesh models of cartoon characters. Using a template-fitting approach, we learn how to adjust an initial mesh to images of a character in various poses. We demonstrate that our model successfully learns to deform the meshes based on the input images. Layering is introduced to handle occlusion and moving limbs. Varying motion and textures are captured with a Deformation Network and an Appearance Refinement Network respectively. We show applications of our model in inbetweening, user-constrained deformation and correspondence estimation.
 In the future, we consider using our model for applications such as motion re-targetting. 

{\small
\bibliographystyle{ieee}
\bibliography{egbib}

\begin{thebibliography}{10}\itemsep=-1pt

\bibitem{AE:2019}
Adobe.
\newblock Adobe after effects, 2019.

\bibitem{Ch:2019}
Adobe.
\newblock Adobe character animator, 2019.

\bibitem{allen2003space}
B.~Allen, B.~Curless, and Z.~Popovi{\'c}.
\newblock The space of human body shapes: reconstruction and parameterization
  from range scans.
\newblock In {\em ACM transactions on graphics (TOG)}, volume~22, pages
  587--594. ACM, 2003.

\bibitem{allen2006learning}
B.~Allen, B.~Curless, Z.~Popovi{\'c}, and A.~Hertzmann.
\newblock Learning a correlated model of identity and pose-dependent body shape
  variation for real-time synthesis.
\newblock In {\em Proceedings of the 2006 ACM SIGGRAPH/Eurographics symposium
  on Computer animation}, pages 147--156. Eurographics Association, 2006.

\bibitem{alp2018densepose}
R.~Alp~G{\"u}ler, N.~Neverova, and I.~Kokkinos.
\newblock Densepose: Dense human pose estimation in the wild.
\newblock In {\em Proceedings of the IEEE Conference on Computer Vision and
  Pattern Recognition}, pages 7297--7306, 2018.

\bibitem{amit1991structural}
Y.~Amit, U.~Grenander, and M.~Piccioni.
\newblock Structural image restoration through deformable templates.
\newblock {\em Journal of the American Statistical Association},
  86(414):376--387, 1991.

\bibitem{arjovsky2017wasserstein}
M.~Arjovsky, S.~Chintala, and L.~Bottou.
\newblock Wasserstein gan.
\newblock {\em arXiv preprint arXiv:1701.07875}, 2017.

\bibitem{bagautdinov2018modeling}
T.~Bagautdinov, C.~Wu, J.~Saragih, P.~Fua, and Y.~Sheikh.
\newblock Modeling facial geometry using compositional vaes.
\newblock {\em In practice}, 1:1, 2018.

\bibitem{baxter2009n}
W.~Baxter, P.~Barla, and K.~Anjyo.
\newblock N-way morphing for 2d animation.
\newblock {\em Computer Animation and Virtual Worlds}, 20(2-3):79--87, 2009.

\bibitem{blanz2003face}
V.~Blanz and T.~Vetter.
\newblock Face recognition based on fitting a 3d morphable model.
\newblock {\em IEEE Transactions on pattern analysis and machine intelligence},
  25(9):1063--1074, 2003.

\bibitem{bousmalis2017unsupervised}
K.~Bousmalis, N.~Silberman, D.~Dohan, D.~Erhan, and D.~Krishnan.
\newblock Unsupervised pixel-level domain adaptation with generative
  adversarial networks.
\newblock In {\em The IEEE Conference on Computer Vision and Pattern
  Recognition (CVPR)}, volume~1, page~7, 2017.

\bibitem{bregler2002mocaptoons}
C.~Bregler, L.~Loeb, E.~Chuang, and H.~Deshpande.
\newblock Turning to the masters: Motion capturing cartoons.
\newblock In {\em Proceedings of the 29th Annual Conference on Computer
  Graphics and Interactive Techniques}, SIGGRAPH '02, pages 399--407, 2002.

\bibitem{catmull1978problems}
E.~Catmull.
\newblock The problems of computer-assisted animation.
\newblock In {\em ACM SIGGRAPH Computer Graphics}, volume~12, pages 348--353.
  ACM, 1978.

\bibitem{chao2010simple}
I.~Chao, U.~Pinkall, P.~Sanan, and P.~Schr{\"o}der.
\newblock A simple geometric model for elastic deformations.
\newblock In {\em ACM transactions on graphics (TOG)}, volume~29, page~38. ACM,
  2010.

\bibitem{choy2016universal}
C.~B. Choy, J.~Gwak, S.~Savarese, and M.~Chandraker.
\newblock Universal correspondence network.
\newblock In {\em Advances in Neural Information Processing Systems}, pages
  2414--2422, 2016.

\bibitem{Correa:1998:TMF}
W.~T. Corr{\^e}a, R.~J. Jensen, C.~E. Thayer, and A.~Finkelstein.
\newblock Texture mapping for cel animation.
\newblock pages 435--446, July 1998.

\bibitem{di2001automatic}
F.~Di~Fiore, P.~Schaeken, K.~Elens, and F.~Van~Reeth.
\newblock Automatic in-betweening in computer assisted animation by exploiting
  2.5 d modelling techniques.
\newblock In {\em Proceedings Computer Animation 2001. Fourteenth Conference on
  Computer Animation (Cat. No. 01TH8596)}, pages 192--200. IEEE, 2001.

\bibitem{dvorovzvnak2017example}
M.~Dvoro{\v{z}}{\v{n}}{\'a}k, P.~B{\'e}nard, P.~Barla, O.~Wang, and
  D.~S{\`y}kora.
\newblock Example-based expressive animation of 2d rigid bodies.
\newblock {\em ACM Transactions on Graphics (TOG)}, 36(4):127, 2017.

\bibitem{dvorovznak2018toonsynth}
M.~Dvoro{\v{z}}n{\'a}k, W.~Li, V.~G. Kim, and D.~S{\`y}kora.
\newblock Toonsynth: example-based synthesis of hand-colored cartoon
  animations.
\newblock {\em ACM Transactions on Graphics (TOG)}, 37(4):167, 2018.

\bibitem{Fan18}
X.~Fan, A.~Bermano, V.~G. Kim, J.~Popovic, and S.~Rusinkiewicz.
\newblock Tooncap: A layered deformable model for capturing poses from cartoon
  characters.
\newblock {\em Expressive}, 2018.

\bibitem{goodfellow2014generative}
I.~Goodfellow, J.~Pouget-Abadie, M.~Mirza, B.~Xu, D.~Warde-Farley, S.~Ozair,
  A.~Courville, and Y.~Bengio.
\newblock Generative adversarial nets.
\newblock In {\em Advances in neural information processing systems}, pages
  2672--2680, 2014.

\bibitem{gregor2013deep}
K.~Gregor, I.~Danihelka, A.~Mnih, C.~Blundell, and D.~Wierstra.
\newblock Deep autoregressive networks.
\newblock {\em arXiv preprint arXiv:1310.8499}, 2013.

\bibitem{groueix2018shape}
T.~Groueix, M.~Fisher, V.~G. Kim, B.~C. Russell, and M.~Aubry.
\newblock Shape correspondences from learnt template-based parametrization.
\newblock {\em arXiv preprint arXiv:1806.05228}, 2018.

\bibitem{gulrajani2017improved}
I.~Gulrajani, F.~Ahmed, M.~Arjovsky, V.~Dumoulin, and A.~C. Courville.
\newblock Improved training of wasserstein gans.
\newblock In {\em Advances in Neural Information Processing Systems}, pages
  5767--5777, 2017.

\bibitem{Hamu2018MultichartGS}
H.~B. Hamu, H.~Maron, I.~Kezurer, G.~Avineri, and Y.~Lipman.
\newblock Multi-chart generative surface modeling.
\newblock {\em SIGGRAPH Asia}, 2018.

\bibitem{henderson2018learning}
P.~Henderson and V.~Ferrari.
\newblock Learning to generate and reconstruct 3d meshes with only 2d
  supervision.
\newblock {\em arXiv preprint arXiv:1807.09259}, 2018.

\bibitem{huang2017stacked}
X.~Huang, Y.~Li, O.~Poursaeed, J.~E. Hopcroft, and S.~J. Belongie.
\newblock Stacked generative adversarial networks.
\newblock In {\em CVPR}, volume~2, page~3, 2017.

\bibitem{isola2017image}
P.~Isola, J.-Y. Zhu, T.~Zhou, and A.~A. Efros.
\newblock Image-to-image translation with conditional adversarial networks.
\newblock {\em arXiv preprint}, 2017.

\bibitem{jacobson2011biharmonic}
A.~Jacobson, I.~Baran, J.~Popovi{\'{c}}, and O.~Sorkine.
\newblock Bounded biharmonic weights for real-time deformation.
\newblock {\em ACM Transactions on Graphics (proceedings of ACM SIGGRAPH)},
  30(4):78:1--78:8, 2011.

\bibitem{kanazawa2018learning}
A.~Kanazawa, S.~Tulsiani, A.~A. Efros, and J.~Malik.
\newblock Learning category-specific mesh reconstruction from image
  collections.
\newblock {\em arXiv preprint arXiv:1803.07549}, 2018.

\bibitem{kato2018neural}
H.~Kato, Y.~Ushiku, and T.~Harada.
\newblock Neural 3d mesh renderer.
\newblock In {\em Proceedings of the IEEE Conference on Computer Vision and
  Pattern Recognition}, pages 3907--3916, 2018.

\bibitem{kiapour2019generating}
M.~Kiapour, S.~Zheng, R.~Piramuthu, and O.~Poursaeed.
\newblock Generating a digital image using a generative adversarial network,
  Sept.~19 2019.
\newblock US Patent App. 15/923,347.

\bibitem{kim2013deformable}
J.~Kim, C.~Liu, F.~Sha, and K.~Grauman.
\newblock Deformable spatial pyramid matching for fast dense correspondences.
\newblock In {\em Proceedings of the IEEE Conference on Computer Vision and
  Pattern Recognition}, pages 2307--2314, 2013.

\bibitem{kingma2014semi}
D.~P. Kingma, S.~Mohamed, D.~J. Rezende, and M.~Welling.
\newblock Semi-supervised learning with deep generative models.
\newblock In {\em Advances in Neural Information Processing Systems}, pages
  3581--3589, 2014.

\bibitem{kingma2013auto}
D.~P. Kingma and M.~Welling.
\newblock Auto-encoding variational bayes.
\newblock {\em arXiv preprint arXiv:1312.6114}, 2013.

\bibitem{kort2002computer}
A.~Kort.
\newblock Computer aided inbetweening.
\newblock In {\em Proceedings of the 2nd international symposium on
  Non-photorealistic animation and rendering}, pages 125--132. ACM, 2002.

\bibitem{kostrikov2018surface}
I.~Kostrikov, Z.~Jiang, D.~Panozzo, D.~Zorin, and B.~Joan.
\newblock Surface networks.
\newblock In {\em 2018 {IEEE} Conference on Computer Vision and Pattern
  Recognition, {CVPR} 2018}, 2018.

\bibitem{ledig2017photo}
C.~Ledig, L.~Theis, F.~Husz{\'a}r, J.~Caballero, A.~Cunningham, A.~Acosta,
  A.~P. Aitken, A.~Tejani, J.~Totz, Z.~Wang, et~al.
\newblock Photo-realistic single image super-resolution using a generative
  adversarial network.
\newblock In {\em CVPR}, volume~2, page~4, 2017.

\bibitem{levi2015smooth}
Z.~Levi and C.~Gotsman.
\newblock Smooth rotation enhanced as-rigid-as-possible mesh animation.
\newblock {\em IEEE transactions on visualization and computer graphics},
  21(2):264--277, 2015.

\bibitem{litany2017deformable}
O.~Litany, A.~Bronstein, M.~Bronstein, and A.~Makadia.
\newblock Deformable shape completion with graph convolutional autoencoders.
\newblock {\em arXiv preprint arXiv:1712.00268}, 2017.

\bibitem{liu2011sift}
C.~Liu, J.~Yuen, and A.~Torralba.
\newblock Sift flow: Dense correspondence across scenes and its applications.
\newblock {\em IEEE transactions on pattern analysis and machine intelligence},
  33(5):978--994, 2011.

\bibitem{loper2015smpl}
M.~Loper, N.~Mahmood, J.~Romero, G.~Pons-Moll, and M.~J. Black.
\newblock Smpl: A skinned multi-person linear model.
\newblock {\em ACM Transactions on Graphics (TOG)}, 34(6):248, 2015.

\bibitem{mirza2014conditional}
M.~Mirza and S.~Osindero.
\newblock Conditional generative adversarial nets.
\newblock {\em arXiv preprint arXiv:1411.1784}, 2014.

\bibitem{oord2016pixel}
A.~v.~d. Oord, N.~Kalchbrenner, and K.~Kavukcuoglu.
\newblock Pixel recurrent neural networks.
\newblock {\em arXiv preprint arXiv:1601.06759}, 2016.

\bibitem{poursaeed2019fine}
O.~Poursaeed, T.~Jiang, H.~Yang, S.~Belongie, and S.-N. Lim.
\newblock Fine-grained synthesis of unrestricted adversarial examples.
\newblock {\em arXiv preprint arXiv:1911.09058}, 2019.

\bibitem{poursaeed2018generative}
O.~Poursaeed, I.~Katsman, B.~Gao, and S.~Belongie.
\newblock Generative adversarial perturbations.
\newblock In {\em Proceedings of the IEEE Conference on Computer Vision and
  Pattern Recognition}, pages 4422--4431, 2018.

\bibitem{radford2015unsupervised}
A.~Radford, L.~Metz, and S.~Chintala.
\newblock Unsupervised representation learning with deep convolutional
  generative adversarial networks.
\newblock {\em arXiv preprint arXiv:1511.06434}, 2015.

\bibitem{ranjan2018generating}
A.~Ranjan, T.~Bolkart, S.~Sanyal, and M.~J. Black.
\newblock Generating 3d faces using convolutional mesh autoencoders.
\newblock {\em arXiv preprint arXiv:1807.10267}, 2018.

\bibitem{revaud2016deepmatching}
J.~Revaud, P.~Weinzaepfel, Z.~Harchaoui, and C.~Schmid.
\newblock Deepmatching: Hierarchical deformable dense matching.
\newblock {\em International Journal of Computer Vision}, 120(3):300--323,
  2016.

\bibitem{rezende2014stochastic}
D.~J. Rezende, S.~Mohamed, and D.~Wierstra.
\newblock Stochastic backpropagation and approximate inference in deep
  generative models.
\newblock {\em arXiv preprint arXiv:1401.4082}, 2014.

\bibitem{salimans2016improved}
T.~Salimans, I.~Goodfellow, W.~Zaremba, V.~Cheung, A.~Radford, and X.~Chen.
\newblock Improved techniques for training gans.
\newblock In {\em Advances in Neural Information Processing Systems}, pages
  2234--2242, 2016.

\bibitem{shewchuk1996triangle}
J.~R. Shewchuk.
\newblock Triangle: Engineering a 2d quality mesh generator and delaunay
  triangulator.
\newblock In {\em Applied computational geometry towards geometric
  engineering}, pages 203--222. Springer, 1996.

\bibitem{shrivastava2017learning}
A.~Shrivastava, T.~Pfister, O.~Tuzel, J.~Susskind, W.~Wang, and R.~Webb.
\newblock Learning from simulated and unsupervised images through adversarial
  training.
\newblock In {\em CVPR}, volume~2, page~5, 2017.

\bibitem{shu2018deforming}
Z.~Shu, M.~Sahasrabudhe, A.~Guler, D.~Samaras, N.~Paragios, and I.~Kokkinos.
\newblock Deforming autoencoders: Unsupervised disentangling of shape and
  appearance.
\newblock {\em arXiv preprint arXiv:1806.06503}, 2018.

\bibitem{soltani2017synthesizing}
A.~A. Soltani, H.~Huang, J.~Wu, T.~D. Kulkarni, and J.~B. Tenenbaum.
\newblock Synthesizing 3d shapes via modeling multi-view depth maps and
  silhouettes with deep generative networks.
\newblock In {\em The IEEE conference on computer vision and pattern
  recognition (CVPR)}, volume~3, page~4, 2017.

\bibitem{sorkine2007rigid}
O.~Sorkine and M.~Alexa.
\newblock As-rigid-as-possible surface modeling.
\newblock In {\em Symposium on Geometry processing}, volume~4, pages 109--116,
  2007.

\bibitem{sun2018pwc}
D.~Sun, X.~Yang, M.-Y. Liu, and J.~Kautz.
\newblock Pwc-net: Cnns for optical flow using pyramid, warping, and cost
  volume.
\newblock In {\em Proceedings of the IEEE Conference on Computer Vision and
  Pattern Recognition}, pages 8934--8943, 2018.

\bibitem{sykora2009rigid}
D.~S{\`y}kora, J.~Dingliana, and S.~Collins.
\newblock As-rigid-as-possible image registration for hand-drawn cartoon
  animations.
\newblock In {\em Proceedings of the 7th International Symposium on
  Non-Photorealistic Animation and Rendering}, pages 25--33. ACM, 2009.

\bibitem{Tan_2018_CVPR}
Q.~Tan, L.~Gao, Y.-K. Lai, and S.~Xia.
\newblock Variational autoencoders for deforming 3d mesh models.
\newblock In {\em The IEEE Conference on Computer Vision and Pattern
  Recognition (CVPR)}, June 2018.

\bibitem{tatarchenko2017octree}
M.~Tatarchenko, A.~Dosovitskiy, and T.~Brox.
\newblock Octree generating networks: Efficient convolutional architectures for
  high-resolution 3d outputs.
\newblock In {\em Proc. of the IEEE International Conf. on Computer Vision
  (ICCV)}, volume~2, page~8, 2017.

\bibitem{van2016conditional}
A.~van~den Oord, N.~Kalchbrenner, L.~Espeholt, O.~Vinyals, A.~Graves, et~al.
\newblock Conditional image generation with pixelcnn decoders.
\newblock In {\em Advances in Neural Information Processing Systems}, pages
  4790--4798, 2016.

\bibitem{venkatdeep}
A.~Venkat, S.~S. Jinka, and A.~Sharma.
\newblock Deep textured 3d reconstruction of human bodies.

\bibitem{wampler2016fast}
K.~Wampler.
\newblock Fast and reliable example-based mesh ik for stylized deformations.
\newblock {\em ACM Transactions on Graphics (TOG)}, 35(6):235, 2016.

\bibitem{whited2010betweenit}
B.~Whited, G.~Noris, M.~Simmons, R.~W. Sumner, M.~Gross, and J.~Rossignac.
\newblock Betweenit: An interactive tool for tight inbetweening.
\newblock In {\em Computer Graphics Forum}, volume~29, pages 605--614. Wiley
  Online Library, 2010.

\bibitem{yang2014daisy}
H.~Yang, W.~Lin, and J.~Lu.
\newblock Daisy filter flow: A generalized approach to discrete dense
  correspondences.
\newblock CVPR, 2014.

\bibitem{yoo2016pixel}
D.~Yoo, N.~Kim, S.~Park, A.~S. Paek, and I.~S. Kweon.
\newblock Pixel-level domain transfer.
\newblock In {\em European Conference on Computer Vision}, pages 517--532.
  Springer, 2016.

\bibitem{zhu2016generative}
J.-Y. Zhu, P.~Kr{\"a}henb{\"u}hl, E.~Shechtman, and A.~A. Efros.
\newblock Generative visual manipulation on the natural image manifold.
\newblock In {\em European Conference on Computer Vision}, pages 597--613.
  Springer, 2016.

\bibitem{zhu2017unpaired}
J.-Y. Zhu, T.~Park, P.~Isola, and A.~A. Efros.
\newblock Unpaired image-to-image translation using cycle-consistent
  adversarial networks.
\newblock {\em arXiv preprint}, 2017.

\bibitem{zuffi2015stitched}
S.~Zuffi and M.~J. Black.
\newblock The stitched puppet: A graphical model of 3d human shape and pose.
\newblock In {\em Proceedings of the IEEE Conference on Computer Vision and
  Pattern Recognition}, pages 3537--3546, 2015.

\end{thebibliography}
}

\end{document}